# Interpretable Deep Learning applied to Plant Stress Phenotyping

**Sambuddha Ghosal**
Department of Mechanical Engineering
Iowa State University
sghosal@iastate.edu

**David Blystone**
Department of Agronomy
Iowa State University
blystone@iastate.edu

**Asheesh K. Singh**
Department of Agronomy
Iowa State University
singhak@iastate.edu

**Baskar Ganapathysubramanian**
Department of Mechanical Engineering
Iowa State University
baskarg@iastate.edu

**Arti Singh**
Department of Agronomy
Iowa State University
arti@iastate.edu

**Soumik Sarkar**
Department of Mechanical Engineering
Iowa State University
soumiks@iastate.edu

## Abstract

Availability of an explainable deep learning model that can be applied to practical real world scenarios and in turn, can consistently, rapidly and accurately identify specific and minute traits in applicable fields of biological sciences, is scarce. Here we consider one such real world example viz., accurate identification, classification and quantification of biotic and abiotic stresses in crop research and production. Up until now, this has been predominantly done manually by visual inspection and require specialized training. However, such techniques are hindered by subjectivity resulting from inter- and intra-rater cognitive variability. Here, we demonstrate a machine learning framework's ability to identify and classify a diverse set of foliar stresses in the soybean plant with remarkable accuracy. We also present an explanation mechanism using gradient-weighted class activation mapping that isolates the visual symptoms used by the model to make predictions. This unsupervised identification of unique visual symptoms for each stress provides a quantitative measure of stress severity, allowing for identification, classification and quantification in one framework. The learnt model appears to be agnostic to species and make good predictions for other (non-soybean) species, demonstrating an ability of transfer learning.

## 1 Introduction

Typical plant stress identification and classification has invariably relied on human experts identifying visual symptoms as a means of categorization [1]. This process is admittedly subjective and error-prone. Computer vision and machine learning have the capability of resolving this issue and enable accurate, scalable high-throughput phenotyping. Among machine learning approaches, deep learning has emerged as one of the most effective techniques in various fields of modern science such as medical imaging applications that have achieved dermatologist level classification accuracies for skin cancer [2], in modeling neural responses and population in visual cortical areas of the brain [3] and in predicting sequence specificities of DNA- and RNA-binding proteins [4]. Similarly, deep learning based techniques have made transformative demonstration in the context of performing complex cognitive tasks such as achieving human level or better accuracy for playing Atari games using Deep Q network [5] and even beating a human expert in playing the Chinese game of Go [6].

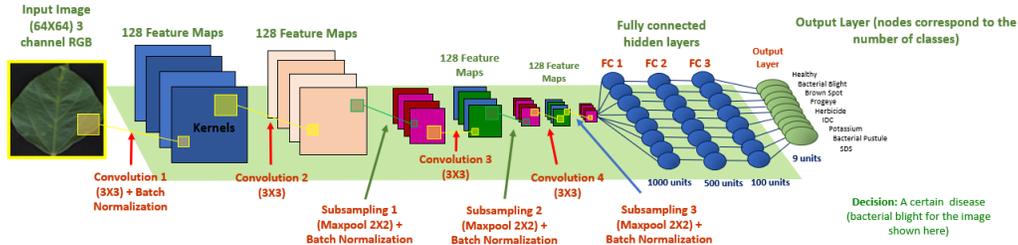

Figure 1: DCNN architecture used.

Figure 2: This table presents leaf image examples for each soybean stress identified and classified by the DCNN model. The Grad-CAM framework was applied to highlight regions of interest (symptoms) extracted by the DCNN model. These automatically extracted symptoms were compared against versions of images with symptoms that were marked manually by expert raters.

We start by building a deep learning model that is exceptionally accurate in identifying a large class base (9 classes) of soybean stresses from RGB images of soybean leaves. However, this type of model typically lack "explainability" which presents a major bottleneck to their widespread acceptance [7]. Here, we sought to explore the trained model to explain each identification and classification decision made by extracting the visual cues or features responsible for a particular decision and determining which region of the leaf image is used by the DCNN model to make a decision and whether this region is correlated with the human-identified symptoms of a particular disease. Our explanation framework is based on concepts of gradient-weighted class activation mapping (Grad-CAM) [8] [9], which queries each prediction of the trained model to extract visual cues (see Figure 1). Figure 2 illustrates the results from this explanation framework with representative examples from each stress. The trained DCNN correctly identified each stress (top row) on the basis of the input image (second row). The explanation framework then (without any supervision) isolated the visual cues (i.e., most important pixels) used by the DCNN for stress identification. These regions are highlighted in red in row 3. With statistical significance, the visual cues identified by the explanation framework correlated with the regions exhibiting visual disease symptoms, as assessed by an expert plant pathologist.

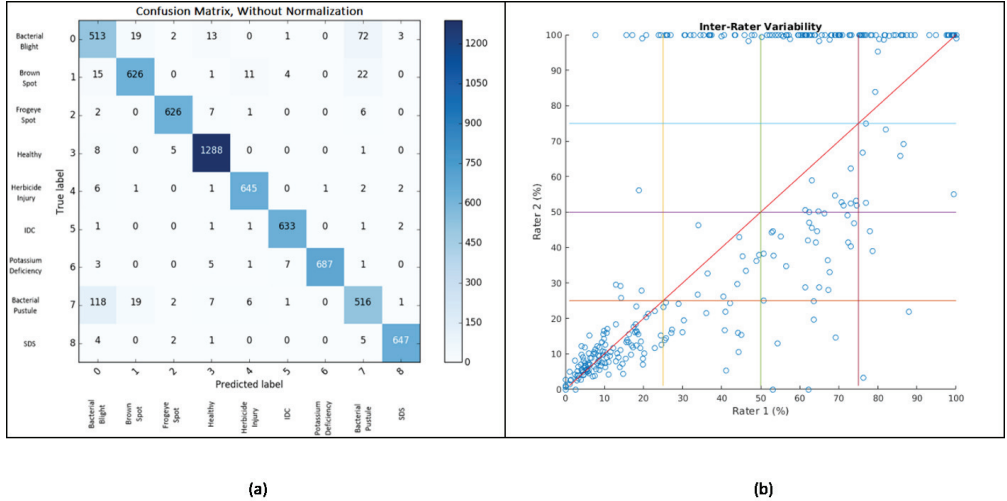

Figure 3: (a) This confusion matrix shows the stress classification results of the DCNN model for eight different stresses and healthy leaves. (b) shows a scatter plot comparing the severity ratings of the same images between two raters for four stresses (IDC, SDS, potassium deficiency and Septoria brown spot) that were pooled and solid red line is the $45^o$ line. The results indicated high inter-rater variability between experienced raters, especially as the stress severity of leaf images increase.

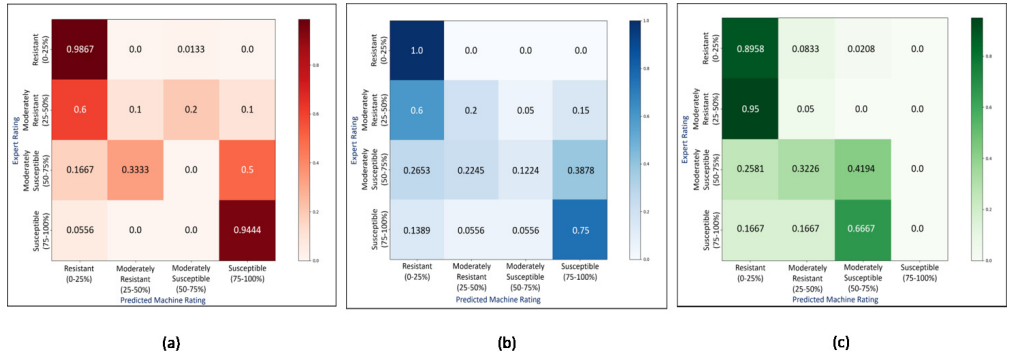

Figure 4: These figures (a, b, c) details the comparison between human and machine learning-based severity ratings for three previously mentioned stresses [(a) Septoria brown spot, (b) IDC, and (c) Sudden Death Syndrome]. The severity comparison using a standard discretized severity scale (0-15%: resistant, 15-30%: moderately resistant, 30-45%: moderately susceptible, 45-75%: susceptible, and 75-100%: highly susceptible) shows the success of the DCNN-based severity estimation framework to correctly quantify symptoms for these stresses.

## 2 The proposed architecture: An explainable Deep Learning Framework

A CNN-based supervised classification framework was developed to identify and classify stresses (Figure 1 (a) and (b)). DCNNs have shown an extraordinary ability [2–6, 10, 11] to efficiently extract complex features from images and function as a classification technique when provided with sufficient data. We associate this classification ability with the hierarchical nature of this model [12], which is able to learn "features of features" from data without the time-consuming hand-crafting of features. We then use the Grad-CAM algorithm [8] to generate heat maps on a test leaf image that signifies the leaf region the DCNN model is focusing on to perform the classification.

### 2.1 The CNN Model

The network architecture (shown in Figure 1) consists of 5 convolutional layers (128 feature maps of size 3X3 for each layer), 4 pooling layers (down sampling by 2X2 max-pooling), 4 batch normal-ization layers and 2 fully connected (FC) layers with 500 and 100 hidden units each, sequentially. Training was performed on a total of 53,265 samples (with an additional 5,919 validation samples), and testing was performed on 6, 576 samples. The learning rate was maintained at 0.1. The Rectified Linear Unit ($ReLU$) function is used as the activation function. To address overfitting issues, we add dropout [13] layers in between the fully-connected (FC) layers. The percentage of dropouts used was 50% after each of the fully-connected layers (namely, FC1 and FC2, as shown in Figure 1). After every convolutional layer, batch normalization was performed to remove internal covariate shift [14]. The network was trained for approximately 100 epochs on the 53,265-image training set to reach the desired accuracy (94%). The crossentropy (categorical) loss (or cost) function along with the Adam optimizer [15] was used to minimize the error.

### 2.2 Gradient-weighted Class Activation Mapping (Grad-CAM)

Grad-CAM increases the transparency of CNN based models and explainability by visualizing the input regions that are more important than others; on this basis, the DCNN then makes predictions. Detailed mathematical formulations and algorithm descriptions can be found in [8]. Grad-CAM follows the CAM approach to localization [16] and enables modification of image classification CNN architectures, in which FC layers are replaced with convolutional layers. Subsequent global-average pooling [17] yields class-specific feature maps. Grad-CAM circumvents issues of CAM by combining feature maps that do not require any modification in the network architecture. This method uses a gradient corresponding to a certain class that is fed into the final convolutional layer of a DCNN to produce an approximate localization (heat) map of the important regions in the image for each class.

## 3 Results

### 3.1 Plant Disease Classification and Identification

Figure 2 presents the qualitative results of deploying the trained DCNN for disease detection and classification. Figure 3 and Figure 4 detail the quantitative results over all test images. We found a high overall classification accuracy (94%) using a large and diverse dataset of unseen test examples (approximately 6000 images, i.e., approximately 600 examples per foliar stress). The confusion matrix revealed that erroneous predictions were predominantly due to confounding disease symptoms that cause confusion even for expert raters (Figure 3b). For example, the highest confusion (17.6% of bacterial pustule (class 7) test images predicted as bacterial blight (class 0) and 11.6% of bacterial blight test images predicted as bacterial pustule) occurred between bacterial blight and bacterial pustule. Discriminating between these two diseases is challenging even for expert plant pathologists [18].

### 3.2 Explaining the symptoms and estimating disease severity

In conventional disease-scouting scenarios, the ratings of even the same expert rater may change depending on various factors (intra-rater variability), such as illumination and human fatigue. Moreover, different human raters often tend to disagree (inter-rater variability), owing to the subjective quantification of the extent of symptoms expressed on a leaf [1, 19]. In contrast, the trained DCNN provides a consistent approach for severity estimation. Specifically, the spatial spread of the automatically identified symptoms allows for estimation of the severity of the classified disease in each leaflet. We computed the severity as the area fraction of identified symptoms and compared it with the severity estimated by an expert human rater. While the algorithm identified symptoms extremely precisely (at a pixel level), the expert human rater estimate was much more qualitative. Instances of inter-rater variability is presented in Figure 3 (b). Figure 4 (a, b, c) shows a comparison of the machine learning-based ratings and human ratings based on a typical discretized severity scale (as mentioned in Figure 4). Furthermore, these rating results demonstrated the efficacy of the DCNN-based severity estimation framework, which identifies the disease symptoms in a completely unsupervised manner. We observed that the few deviations in these results were primarily due to the low quality of the corresponding images, which exhibited shadows, low resolution and a lack of focus.

# References

[1] Bock C., Poole G., Parker P. & Gottwald T. (2010) Plant disease severity estimated visually, by digital photography and image analysis, and by hyperspectral imaging. Critical Reviews in Plant Sciences 29(2):59–107.

[2] Esteva A., et al. (2017) Dermatologist-level classification of skin cancer with deep neural networks. Nature 542(7639):115–118.

[3] Yamins D.L. & DiCarlo J.J. (2016) Using goal-driven deep learning models to understand sensory cortex. Nature neuroscience 19(3):356.

[4] Alipanahi B., Delong A., Weirauch M.T. & Frey BJ. (2015) Predicting the sequence specificities of dna-and rna-binding proteins by deep learning. Nature biotechnology 33(8):831–838.

[5] Mnih V., et al. (2015) Human-level control through deep reinforcement learning. Nature 518(7540):529–533.

[6] Silver D., et al. (2016) Mastering the game of go with deep neural networks and tree search. Nature 529(7587):484–489.

[7] Castelvecchi D. (2016) Can we open the black box of ai? Nature News 538(7623):20.

[8] Selvaraju R.R., et al. (2016) Grad-cam: Visual explanations from deep networks via gradientbased localization. See https://arxiv. org/abs/1610.02391 v3.

[9] Balu A, et al. (2017) Learning localized geometric features using 3d-cnn: An application to manufacturability analysis of drilled holes.

[10] Sladojevic S., Arsenovic M., Anderla A., Culibrk D. & Stefanovic D. (2016) Deep neural networks based recognition of plant diseases by leaf image classification. Computational intelligence and neuroscience 2016.

[11] Ubbens J.R. & Stavness I. (2017) Deep plant phenomics: A deep learning platform for complex plant phenotyping tasks. Frontiers in plant science 8.

[12] Stoecklein D., Lore K.G., Davies M., Sarkar S. & Ganapathysubramanian B. (2017) Deep learning for flow sculpting: Insights into efficient learning using scientific simulation data. Scientific Reports 7:srep46368.

[13] Srivastava N., Hinton G.E., Krizhevsky A., Sutskever I. & Salakhutdinov R. (2014) Dropout: a simple way to prevent neural networks from overfitting. Journal of machine learning research 15(1):1929–1958.

[14] Ioffe S. & Szegedy C. (2015) Batch normalization: Accelerating deep network training by reducing internal covariate shift in International Conference on Machine Learning. pp. 448–456.

[15] Kingma D. & Ba J. (2014) Adam: A method for stochastic optimization. arXiv preprint arXiv:1412.6980.

[16] Zhou B., Khosla A., Lapedriza A., Oliva A. & Torralba A. (2016) Learning deep features for discriminative localization in Proceedings of the IEEE Conference on Computer Vision and Pattern Recognition. pp. 2921–2929.

[17] Lin M., Chen Q. & Yan S. (2013) Network in network. arXiv preprint arXiv:1312.4400.

[18] Hartman G.L., et al. (2015) Compendium of soybean diseases and pests. (Am Phytopath Society)

[19] Chiang K.S., Bock C.H., Lee I.H., El Jarroudi M. & Delfosse P. (2016) Plant disease severity assessment — how rater bias, assessment method, and experimental design affect hypothesis testing and resource use efficiency. Phytopathology 106(12):1451–1464.